\documentclass[10pt,journal,compsoc]{IEEEtran}
\ifCLASSOPTIONcompsoc
  \usepackage[nocompress]{cite}
\else
  \usepackage{cite}
\fi
\ifCLASSINFOpdf
\else
\fi

\usepackage{hyperref}
\usepackage{url}

\usepackage{microtype}
\usepackage{graphicx}
\usepackage{subfig}
\usepackage{booktabs} 
\usepackage{dblfloatfix} 
\usepackage{hyperref}

\usepackage{amsmath,amssymb,amsfonts}
\usepackage{algorithmic}
\usepackage{algorithm}
\usepackage{graphicx}
\usepackage{textcomp}
\usepackage{xcolor}
\usepackage{enumitem}

\usepackage{amsthm}
\usepackage{tikz,pgfplots}
\usetikzlibrary{shadows}
\usetikzlibrary{arrows,shapes.geometric}
\usetikzlibrary{patterns}
\usepackage{caption}
\definecolor{rosso}{RGB}{220,57,18}
\definecolor{giallo}{RGB}{255,153,0}
\definecolor{blu}{RGB}{102,140,217}
\definecolor{verde}{RGB}{16,150,24}
\definecolor{viola}{RGB}{153,0,153}
\usepackage{fix-cm}

\usepackage{url}
\usepackage{booktabs} 
\usepackage{makecell}
\usepackage{multirow}
\usepackage{multicol}
\usepackage{graphicx}
\graphicspath{{img/}}

\usepackage{amssymb,xpatch}
\newtheorem{prop}{Proposition}
\usepackage{amssymb}
\usepackage{adjustbox}

\usepackage{thmtools}
\usepackage{float}
\usepackage{balance}

\usepackage{colortbl}
\usepackage{flushend}

\usepackage{cellspace}%
\usepackage{makecell}
\setcellgapes{3pt}

\tikzset{
  chart/.style={
    legend label/.style={font={\scriptsize},anchor=west,align=left},
    legend box/.style={rectangle, draw, minimum size=5pt},
    axis/.style={black,semithick,->},
    axis label/.style={anchor=west,font={\tiny}},
  },
  bar chart/.style={
    chart,
    bar width/.code={
        \pgfmathparse{##1/2}
        \global\let\bar@w\pgfmathresult
    },
    bar/.style={very thick, draw=white},
    bar label/.style={font={\bfseries\small},anchor=north},
    bar value/.style={font={\footnotesize}},
    bar width=.75,
  },
  pie chart/.style={
    chart,
    slice/.style={line cap=round, line join=round, thin,draw=white},
    pie title/.style={font={\small}},
    slice type/.style n args={3}{
        ##1/.style={pattern color=##2,pattern=##3},
        values of ##1/.style={font=\fontsize{0.1}{4}\selectfont, color = white}
    }
}}

\pgfdeclarelayer{background}
\pgfdeclarelayer{foreground}
\pgfsetlayers{background,main,foreground}

\newcommand{\pie}[3][]{
    \begin{scope}[#1]
    \pgfmathsetmacro{\curA}{90}
    \pgfmathsetmacro{\r}{1}
    \def\c{(0,0)}
    \node[pie title] at (90:1.3) {#2};
    \foreach \v/\s in{#3}{
        \pgfmathsetmacro{\deltaA}{\v/100*360}
        \pgfmathsetmacro{\nextA}{\curA + \deltaA}
        \pgfmathsetmacro{\midA}{(\curA+\nextA)/2}
        \path[slice,\s] \c
            -- +(\curA:\r)
            arc (\curA:\nextA:\r)
            -- cycle;
        \pgfmathsetmacro{\d}{max((\deltaA * -(.5/50) + 1) , .5)}

        \begin{pgfonlayer}{foreground}
        \path \c -- node[pos=\d,pie values,values of \s]{$\v\%$} +(\midA:\r);
        \end{pgfonlayer}
        \global\let\curA\nextA
    }
    \end{scope}
}

\newcommand{\legend}[2][]{
    \begin{scope}[#1]
    \path
        \foreach \n/\s in {#2}
            {
                  ++(0,-10pt) node[\s,legend box] {} +(5pt,0) node[legend label] {\n}
            }
    ;
    \end{scope}
}

\usepackage{mathtools}  
\usepackage{amssymb}  
\usepackage{enumitem}

\pgfplotsset{
  /pgfplots/xlabel near ticks/.style={
     /pgfplots/every axis x label/.style={
        at={(ticklabel cs:0.5)},anchor=near ticklabel
     }
  },
  /pgfplots/ylabel near ticks/.style={
     /pgfplots/every axis y label/.style={
        at={(ticklabel cs:0.5)},rotate=90,anchor=near ticklabel}
     }
  }
\makeatletter
\let\oldtagform@\tagform@
\renewcommand{\eqref}[1]{\textup{\oldtagform@{\ref{#1}}}}
\makeatother

\usepackage{amsmath}
\usepackage{mathtools,amssymb}

\usepackage{soul}

\begin{document}


%
\title{Improving Mutual Information based Feature Selection by Boosting Unique Relevance}
%
%
%
%

\author{Shiyu~Liu,~\IEEEmembership{Student Member,~IEEE,}
        Mehul~Motani,~\IEEEmembership{Fellow,~IEEE}
        
\IEEEcompsocitemizethanks{\IEEEcompsocthanksitem The authors are with the Department of Electrical and Computing Engineering, National University of Singapore, 4 Engineering Drive 3, Singapore 117853 \protect\\
E-mail: shiyu\_liu@u.nus.edu, motani@nus.edu.sg}}

\IEEEtitleabstractindextext{%
\begin{abstract}
Mutual Information (MI) based feature selection makes use of MI to evaluate each feature and eventually shortlists a relevant feature subset, in order to address issues associated with high-dimensional datasets. Despite the effectiveness of MI in feature selection, we notice that many state-of-the-art algorithms disregard the so-called unique relevance (UR) of features, which is a necessary condition for the optimal feature subset. In fact, in our study of five representative MI based feature selection (MIBFS) algorithms, we find that all of them underperform as they ignore the UR of features and arrive at a suboptimal selected feature subset which contains a non-negligible number of redundant features. We point out that the heart of the problem is that all these MIBFS algorithms follow the criterion of Maximize Relevance with Minimum Redundancy (MRwMR), which does not explicitly target UR. This motivates us to augment the existing criterion with the objective of boosting unique relevance (BUR), leading to a new criterion called MRwMR-BUR. Depending on the task being addressed, MRwMR-BUR has two variants, termed MRwMR-BUR-KSG and MRwMR-BUR-CLF, which estimate UR differently. MRwMR-BUR-KSG estimates UR via a nearest-neighbor based approach called the KSG estimator and is designed for three major tasks: (i) Classification Performance (i.e., higher classification accuracy). (ii) Feature Interpretability (i.e., a more precise selected feature subset for practitioners to explore the hidden relationship between features and labels). (iii) Classifier Generalization (i.e., the selected feature subset generalizes well to various classifiers). MRwMR-BUR-CLF estimates UR via a classifier based approach. It adapts UR to different classifiers, further improving the competitiveness of MRwMR-BUR for classification performance oriented tasks. The performance of both MRwMR-BUR-KSG and MRwMR-BUR-CLF is validated via experiments using six public datasets and three popular classifiers. Specifically, as compared to MRwMR, the proposed MRwMR-BUR-KSG improves the test accuracy by 2\% - 3\% with 25\% - 30\% fewer features being selected, without increasing the algorithm complexity. MRwMR-BUR-CLF further improves the classification performance by 3.8\% - 5.5\% (relative to MRwMR), and it also outperforms three popular classifier dependent feature selection methods.
\end{abstract}

}

\maketitle

\IEEEdisplaynontitleabstractindextext

%
\IEEEpeerreviewmaketitle

\IEEEraisesectionheading{\section{Introduction}\label{sec:introduction}}

\IEEEPARstart{H}{igh}-dimensional datasets tend to contain irrelevant and redundant features, leading to extra computation, larger storage, and degraded performance \cite{Bengio2013,Gao2016, el2020improved,zhou2022feature,salem2022fuzzy}. Mutual Information (MI) \cite{cover2006elements} based feature selection, which is a classifier independent filter method in the field of dimensionality reduction, attempts to address those issues by selecting a relevant feature subset. We start this work by discussing the advantage of MI based feature selection (MIBFS) over other types of dimensionality reduction methods.

{\bf (1) Feature Interpretability.} Dimensionality reduction methods can be divided into two classes: feature extraction and feature selection. Feature extraction transforms original features into new features with lower dimensionality while preserving the key information in the original features. For example, the principal component analysis \cite{jolliffe2005principal} projects data points onto only the first few principal components while preserving as much of the data's variation as possible. Feature extraction may perform well in dimensionality reduction, but the extraction process (e.g., projection) loses the physical meaning of features \cite{G2014,Sun2014,Nguyen2014,lamba2022hybrid}. In contrast, feature selection preserves the feature interpretability by selecting a relevant feature subset. This helps to 
explore the hidden relationship between variables and makes techniques such as MIBFS preferred in various domains \cite{Been2015,G2014,hassan2022epileptic,tripathi2020interpretable}. Examples of such studies are \cite{el2020novel, sun2020adaptive, too2021hyper}, which make use of feature selection to shortlist a set of medical features associated with the coronavirus. Studying the selected feature subset could provide new insights on the diagnosis of coronavirus and further advance the research on it.

{\bf (2) Classifier Generalization.} Feature selection methods are either classifier dependent or classifier independent \cite{Guyon2003,G2014}. Examples of the former type include the wrapper method (e.g., forward feature selection \cite{marcano2010feature}) and the embedded method which performs feature selection during the training of a pre-defined classifier (e.g., LASSO with the L$1$ penalty \cite{LASSO}). The classifier dependent method tends to provide good performance as it directly makes use of the interaction between features and accuracy. However, the selected features are optimized for the pre-defined classifier and may not perform well for other classifiers \cite{solorio2020review, zebari2020comprehensive,bommert2020benchmark}. The filter method, which is classifier independent, scores each feature according to its relevance with the label. 
As a filter method, MIBFS quantifies relevance using MI as MI can capture the dependencies between random variables (i.e., feature and label). Consequently, the feature subset selected by MIBFS is not tied to the bias of the classifier and is easier to generalize to different classifiers \cite{Bengio2013,Meyer2008,cai2018feature,khaire2019stability,gu2022feature}.

{\bf (3) Classification Performance.} The objective of MIBFS is to find  the minimal feature subset with maximum MI with respect to the label \cite{Brown}. Mathematically, the goal can be written as
\begin{equation} 
\label{NewP}
S^* = \mathop{\arg\min} \\ f(\mathop{\arg\max}_{S\subseteq\Omega} I (S;Y)),
\end{equation}
where $f (A,B,\cdots) = (|A|,|B|,\cdots)$, $|A|$ represents the number of features in $A$ and $\Omega$ is the set of all features, $S \subseteq \Omega$ is the selected feature subset and $S^*$ is the optimal feature subset. Several works \cite{shiyuISIT, Brown} have proved that maximizing  $I (S;Y)$ is equivalent to maximizing the likelihood ($p(Y|S)$) and suggest that minimizing the size of $S$ helps to improve the generalization performance on the unseen data (more details in Sections \ref{sec3.1} and \ref{sec3.2}).
Finding the optimal feature subset through exhaustive search is computationally intractable. Therefore, numerous MIBFS algorithms \cite{Meyer2008,Yang1999,Nguyen2014,Peng2005} attempt to select the optimal feature subset following the criterion of Maximize Relevance with Minimum Redundancy (MRwMR) \cite{Peng2005}, where Maximize Relevance corresponds to the requirement of maximizing $I(S;Y)$ and Minimum Redundancy is to reduce the size of $S$. Those algorithms have provided competitive performance in dimensionality reduction (see recent survey works \cite{zebari2020comprehensive,venkatesh2019review}).

In this paper, we explore a promising feature property, called Unique Relevance (UR), which is the key to select the optimal feature subset in \eqref{NewP}. We note that UR has been defined for a long time and it is also known as strong relevance \cite{KJ}. However, only very few works \cite{shiyuBIBM,shiyuISIT} look into it and the use of UR for feature selection remains largely unexplored.
We fill in this gap and improve the performance of MIBFS by exploring the utility of UR. We summarize the flow of the remaining paper together with our contributions to the field of MIBFS as follows.
\begin{enumerate}[noitemsep,leftmargin=5mm, topsep=-0pt]
\item We shortlist five representative MIBFS algorithms and uncover the fact that all of them ignore UR and end up underperforming, namely they select a non-negligible number of redundant features, contradicting the objective of minimal feature subset in \eqref{NewP} (see Section \ref{sec3.4}). 

\item We point out that the heart of the problem is that existing MIBFS algorithms follow the criterion of MRwMR \cite{Peng2005}, which lacks a mechanism to identify the UR of features. This motivates us to augment MRwMR and include the objective of boosting UR, leading to a new criterion for MIBFS, called MRwMR-BUR (see Section \ref{sec3.5}).

\item We estimate UR via a nearest neighbor based approach called KSG estimator \cite{Kraskov2003}, resulting in the first variant of MRwMR-BUR, called MRwMR-BUR-KSG. The MRwMR-BUR-KSG is designed to improve MRwMR for three major tasks: (i) Classification Performance (an improvement of 2 - 3\% in test accuracy). (ii) Feature Interpretability (i.e., 25 - 30\% fewer unnecessary features are selected). (iii) Classifier Generalization (i.e., the selected feature subset generalizes well to all classifiers studied) (see Section \ref{sec4}).



\item We propose a classifier based approach to estimate UR, resulting in the second variant of MRwMR-BUR, called MRwMR-BUR-CLF. The MRwMR-BUR-CLF adapts UR to different classifiers, further improving the competitiveness of MRwMR-BUR for classification performance oriented tasks. Through extensive experiments, we observe that MRwMR-BUR-CLF further improves the classification performance of MRwMR by 3.8\% - 5.5\% and and it also outperforms three popular classifier dependent feature selection methods (see Section \ref{sec5}).


\end{enumerate}

We note that a short version of this work has been published in \cite{shiyuISIT}, and this work extends \cite{shiyuISIT} in three aspects: (i) We evaluate the proposed MRwMR-BUR criterion from the perspective of achieving the goal of MIBFS stated in \eqref{NewP}. Our experimental results in Section \ref{sec3.5} demonstrate that the proposed MRwMR-BUR can better achieve the goal of MIBFS than MRwMR. This serves as a more solid motivation for the proposed MRwMR-BUR. (ii) In addition to the classification accuracy, the performance of MRwMR-BUR is also evaluated in terms of Feature Interpretability (i.e., if the selected feature subset is precise for practitioners to explore the hidden relationship between features and labels) and Classifier Generalization (i.e., if the selected feature subset generalizes well to various classifiers). (iii) We improve the competitiveness of MRwMR-BUR for performance oriented tasks and propose a classifier based approach to estimate UR, leading to another variant of MRwMR-BUR, called MRwMR-BUR-CLF. The MRwMR-BUR-CLF further improves the performance of MRwMR by 3.8\% - 5.5\% and it also outperforms three popular classifier based feature selection methods.

\section{Background}
\label{BD}
We first provide a brief introduction to information theoretic concepts in Section \ref{sec2.2}, followed by a definition of the notation in Section \ref{sec2.1}. Next, in Section \ref{sec2.3}, we review existing works and shortlist five MRwMR based algorithms. In Section \ref{sec2.4}, we present three types of information content (unique relevance, conditional relevance, irrelevance). In Section \ref{EMI}, we have a discussion on the estimation of MI.

\subsection{Entropy and Mutual Information}
\label{sec2.2}
\textbf{(1) Entropy.} Let $X$ be a discrete random variable with alphabet $\mathcal{X}$ and probability mass function $p(x) = Pr(X = x), x \in \mathcal{X} $. The entropy $H(X)$ measures the uncertainty present in the distribution of $X$ \cite{cover2006elements} and is defined as 

\begin{equation}
H (X)  = - \underset{x \in \mathcal{X}}{\sum} p(x) \log p(x).
\end{equation}

To compute it, one possible approach is to estimate the distribution of $p$($X$) by the frequency counts from data (i.e., the fraction of observations taking on value $x$ from the total $N$). We will discuss more details in Section \ref{EMI}. If the distribution of $X$ is highly biased toward one particular event $x \in \mathcal{X} $, then the entropy is low. This makes sense as we have less uncertainty about the outcome. If all events are equally likely, we have maximum uncertainty over the outcome, then $H(X)$ is maximal.

\textbf{(2) Conditional Entropy.} The conditional entropy $H(Y|X)$ of a pair of discrete random variables $(X, Y)$ with a joint distribution $p(x,y)$ is defined as
\begin{align}
H(Y|X) & =- \underset{x \in \mathcal{X}}{\sum} p(x) H(Y|x) \\
& = - \underset{x \in \mathcal{X}}{\sum} \underset{y \in \mathcal{Y}}{\sum} p(x, y) \log p(y|x).
\end{align}
This can be thought of as the amount of uncertainty remained in $Y$ after learning the outcome of $X$.

\textbf{(3) Mutual Information.} The mutual information \cite{cover2006elements} between two discrete random variables $X$ and $Y$ with a joint probability mass function $p(x, y)$ and marginal probability mass functions $p(x)$ and $p(y)$ is defined as
\begin{align}
\label{mieq}
I(X;Y) = \underset{x \in \mathcal{X}}{\sum} \underset{y \in \mathcal{Y}}{\sum}  p(x, y) \log \frac{p(x,y)}{p(x)p(y)}.
\end{align}
By chain rule for mutual information, \eqref{mieq} can be rewritten as
\begin{equation}
I (X;Y)= H(Y) - H(Y|X).
\end{equation}
This can be interpreted as the reduction in the uncertainty of $Y$ due to the knowledge of $X$. Intuitively, it is the amount of information that one variable provides about another one. We note that the mutual information is symmetric (i.e., $I(X; Y) = I(Y; X)$), and is zero if and only if the variables are statistically independent. In contrast to correlation, MI captures non-linear relationship between variables, and thus can act as a measure of true dependence.

\textbf{(4) Conditional Mutual Information.} The conditioned form of mutual information is given by
\begin{align}
I(X;Y|Z) &= H(X|Z) - H(X|YZ) \\
&= \underset{z \in \mathcal{Z}}{\sum} p(z) \underset{x \in \mathcal{X}}{\sum} \underset{y \in \mathcal{Y}}{\sum} p(xy|z) \log \frac{p(xy|z)}{p(x|z)p(y|z)}.
\end{align}
The conditional mutual information captures the information still shared between $X$ and $Y$ after knowing $Z$, which is a very important property to be considered in feature selection.

\subsection{Notation}
\label{sec2.1}
We now define the notation used in this paper.
We denote the set of all features by $\Omega = \{X_k, k=1,\cdots,M\}$, where $M$ is the number of features. The feature $X_k\in\Omega$ and the label $Y$ are both vectors of length $N$, where $N$ is the number of samples. Let $S \subseteq \Omega$ be the set of selected features and $\tilde{S} \subseteq \Omega$ be the set of unselected features, i.e., $\Omega = \{ S, \tilde{S} \}$.

\subsection{Prior MIBFS Works}
\label{sec2.3}
In MIBFS, MI and its variants (e.g., conditional MI, joint MI) are used as the core of a scoring function to measure how potentially useful a feature or a feature subset could be. Generally, the scoring function will assign a score to each unselected feature and sequentially selects the feature with the maximum score in the current iteration. The procedure for MIBFS using the scoring function $J_{ABC} (\cdot)$ is summarized in Algorithm \ref{algorithm1}. We now describe the scoring functions of five representative MIBFS algorithms as follows.

The Mutual Information Maximization (MIM) 
\cite{Lewis1992} was the first work to use MI as a scoring function. It scores each feature independently of the others, which is simple but a limitation in itself. The scoring function is given by
\begin{equation} \label{eq6}
J_{\text{MIM}}(X_k) = I (X_k ;Y).
\end{equation}

The minimal Redundancy Maximum Relevance (mRMR) \cite{Peng2005} introduced an additional penalty term to reduce redundancy within the selected feature set $S$. The term is divided over the cardinality of $S$ to adaptively vary the penalty. The scoring function is given by
\begin{equation}
J_{\text{mRMR}}(X_k) = I(X_k;Y) - \frac{1}{|S|} \underset{X_j \in S}{\sum} I(X_k, X_j).
\end{equation}

The Joint Mutual Information (JMI) \cite{Yang1999,Meyer2008} was proposed to increase complementary information between features. The scoring function is given by
\begin{equation} \label{eq7}
J_{\text{JMI}}(X_k) = \displaystyle\sum_{X_{j}\in\mathcal{S}}^{}I (X_k,X_{j} ;Y).
\end{equation}

The Joint Mutual Information Maximization (JMIM) \cite{Bennasar2015} built upon JMI and attempted to alleivate the problem of overestimation of the feature significance, so as to optimize the relationship between relevance and redundancy. The scoring function is given by
\begin{equation} \label{eq9}
J_{\text{JMIM}}(X_k)= \underset{X_{j}\in\mathcal{S}}{\overset{}{\text{min}}}I(X_k,X_{j};Y).
\end{equation}

The Greedy Search Algorithm (GSA) \cite{Brown} is a forward searching method that works by selecting, at each step, the candidate feature with the largest joint MI with the label. The scoring function is given by
\begin{equation} \label{eq11}
J_{\text{GSA}}(X_k) = I(X_k, S;Y).
\end{equation}

Along the way, many interesting works have proposed to improve the performance of MIBFS. As an example, \cite{wang2015feature} made use of conditional MI and attempted to minimize the global redundancy (i.e., the overall redundancy within the selected feature subset). \cite{7812571} proposed a MI based term called independent classification information, aiming to strike a balance between redundancy and relevance. \cite{zhang2018muse} introduced a term called uncertainty score, evaluated by conditional entropy, and this term is minimized during feature selection, leading to a method called minimum uncertainty and feature sample elimination. More recently, \cite{gao2020feature} penalized the relevance between a candidate feature and the selected feature within the same class, leading to a method called min-redundancy and max-dependency. \cite{song2021feature} proposed bar bones particle swarm optimization to effectively estimate the value of MI during feature selection.

Overall, all these heuristics aim to find the optimal feature subset $S^*$ in \eqref{NewP} via the criterion of Maximize Relevance with Minimum Redundancy \cite{Peng2005}, where Maximize Relevance corresponds to the requirement of maximizing $I(S;Y)$ and Minimum Redundancy is to reduce the size of $S$.

\begin{algorithm}[!t]
\caption{MI based Feature Selection via $J_{ABC}$}
\label{algorithm1}
\begin{enumerate}
	\itemsep0em 
	\item Input: Scoring function $J_{ABC}(\cdot)$; \\ $\Omega$ $\leftarrow$ Set of $M$ features; \\ $Y\leftarrow$ Label;
	\item Initialization:
	$S$ $\leftarrow \left\{\emptyset\right\}$; $K\le M$.
	\item Repeat until $|S| = K$:  \\
	\hspace*{2mm} Choose the feature $F$ that
    $F=\underset{X\in\Omega\backslash S}{\overset{}{\text{~argmax~}}} J_{ABC}(X);$\\
	\hspace*{2mm} $S \leftarrow S \cup {F} $;
	\item Output the set $S$ with the selected features.
\end{enumerate}
\end{algorithm}


\subsection{Information Content: UR, CR, Irrelevance}
\label{sec2.4}
In feature selection, a feature may contain three types of information with respected to the label. They are Unique Relevance (UR), Conditional Relevance (CR) and Irrelevance.
 
The Unique Relevance (UR) of a feature $X_k$ is defined as the unique relevant information which is not shared by any other features in $\Omega$. Mathematically, UR can be calculated as the MI loss when removing that feature from $\Omega$. By the chain rule for MI \cite{cover2006elements}, UR can be written as
\begin{equation} \label{URE}
\text{UR ($X_k$)} =I(\Omega;Y) - I(\Omega \backslash X_k;Y) = I(X_k;Y|\Omega \backslash X_k).
\end{equation}
Features with non-zero UR are called unique relevant features (also known as strong relevant features in \cite{KJ,Brown}). 

The Conditional Relevance (CR) of a feature $X_k$ to the label $Y$ with respect to a feature set $W$ can be written as 
\begin{equation}
\text{CR} = I(X_k;Y|W).
\end{equation}
Features with non-zero CR are called conditionally relevant features while features with zero CR are called conditional irrelevant features \cite{Brown, yu2004efficient}. We note that there is another definition called weak relevant feature in the literature \cite{KJ,Brown}. A conditional relevant feature $X_k$ (i.e., $I(X_k;Y|W) > 0$) is equivalent to the weak relevant feature if it satisfies two conditions: (i) feature $X_k$ has zero UR. (ii) $W \subseteq \{ \Omega \backslash X_k \}$.  

Irrelevance can be understood as the noise in the signal. Overfitting to the irrelevant aspects of the data will confuse the classifier, leading to decreased accuracy \cite{john1994, song2011}. Mathematically, we define irrelevance of feature $X_k$ as follows,
\begin{equation}
\text{Irrelevance ($X_k$)}=H(X_k) - I(X_k;Y) = H(X_k|Y). \label{irr}
\end{equation}
We note that a feature $X_k$ can be completely irrelevant with respect to the label Y if $I(X_k;Y)$ = 0.

There is another popular type of decomposition called partial information decomposition (PID) \cite{williams2010nonnegative} which decomposes the total mutual information of a system into three parts: unique information, redundancy, synergy and a follow-up work \cite{bertschinger2014quantifying} attempts to quantify each term based on ideas from decision theory. We note that the definition of UR is similar to unique information, but is estimated differently.

\subsection{Estimation of Mutual Information}
\label{EMI}
MI is considered to be very powerful and, therefore, has inspired many works on its estimation. There are two basic approaches to estimate MI -- parametric and non-parametric.

The parametric approach is a given form for the density function which assumes that the data are from a known family of distributions (e.g., Gaussian) and the parameters of the function (i.e., mean and variance) are then optimized by fitting the model to the data set \cite{walters2009estimation, batina2011mutual}. As an example, \cite{belghazi2018} proposed Mutual Information Neural Estimation (MINE) based on dual representations of the KL-divergence and optimize its parameters via back-propagation.
More recently, \cite{rMINE} introduced a regularized version of MINE which aims to enhance the stability of the original MINE while \cite{9518097} attempted to improve MINE with the label smoothing.

The non-parametric approach utlizes the  geometry of the underlying sample to
estimate the local density. Examples include histogram-based estimator \cite{moddemeijer1999statistic, walters2009estimation}, adaptive partitioning \cite{darbellay1999estimation}, kernel density estimator \cite{batina2011mutual} and $K$ nearest neighbor based estimator \cite{Kraskov2003}. Compared to the parametric approach, the non-parametric approach makes no assumption about the distribution of data and hence, is more practical.

In this paper, we estimate MI using the KSG estimator \cite{Kraskov2003} which uses the $K$ nearest neighbors of points in the dataset to detect structure in the underlying probability distribution. Compared to other non-parametric approaches, the KSG estimator is more computationally efficient, and therefore it is more suitable for intensive estimation of MI in MIBFS. Furthermore, the nearest neighbor based approach is more robust to noise and in a recent work \cite{gao2018demystifying}, the KSG estimator is proven to be consistent under some mild assumptions.
It is worth noting that the KSG estimator is not applicable when the random variable being studied is a mixture of continuous and discrete values. 
For that case, we can apply the mixed KSG estimator \cite{gao2017estimating}, which demonstrates good performance at handling mixed variables. We note that the features of all datasets studied in this paper are either purely discrete (real-valued) or continuous while all labels are purely discrete (real-valued) (see Table \ref{tab:rf} rows 1 - 5). Therefore, we use the KSG estimator to compute MI quantities. 

\section{A New Criterion for MIBFS}
\label{sec3}
In this section, we first decompose the goal of MIBFS (stated in \eqref{NewP}) into two parts: (i) Why Maximum MI?; (ii) Why Minimal Feature Subset? and answer these two questions in Section \ref{sec3.1} and Section \ref{sec3.2}, respectively. In Section \ref{sec3.3}, we show the crucial role of UR in selecting the optimal feature subset in \eqref{NewP}. Next, in Section \ref{sec3.4}, we conduct experiments to uncover the fact that all studied MIBFS algorithms are underperforming in achieving the goal of MIBFS. Lastly, in Section \ref{sec3.5}, we motivate MRwMR-BUR as a new criterion for MIBFS and demonstrate that MRwMR-BUR could better achieve the goal of MIBFS.

\subsection{Goal of MIBFS: Why Maximum MI?}
\label{sec3.1}
We denote the dataset by $D$ = $\{ (\Omega^i,Y^i): i=1..N \}$, where $\Omega^i$, $Y^i$, $S^i$ denote all features, label and selected features for the $i_{th}$ sample, respectively. Recall the goal of MIBFS is to identify the minimal feature subset $S$  to maximize the MI with the label. The latter part of the goal can be written as 
\begin{align}
l^*&=\max_{S\subseteq\Omega}\; I(S;Y) . \label{MI_1}
\end{align}
By chain rule for MI, \eqref{MI_1} can be rewritten as
\begin{align}
l^*&= \max_{S\subseteq\Omega}\; I(S;Y) = \max_{S\subseteq\Omega}\; H(Y) - H(Y|S). \label{MI_2}
\end{align}
The H(Y) is constant as it quantifies the uncertainty in Y and is not going to change during feature selection. Therefore, \eqref{MI_2} can further rewritten as
\begin{align}
l^*& = \max_{S\subseteq\Omega}\; -H(Y|S) = \max_{S\subseteq\Omega}\; \mathbb{E}_{sy} \left\{ \log{p(Y|S)} \right\}.  \label{MI_3} 
\end{align}
By the definition of conditional entropy, \eqref{MI_3} can be approximated as 
\begin{align}
l^*& = \max_{S\subseteq\Omega}\; \mathbb{E}_{sy} \left\{ \log{p(Y|S)} \right\} \nonumber  \\
& \approx \max_{S\subseteq\Omega}\frac{1}{N} \sum_{i=1}^{N} \log{p(y^i|S^i)} = \max_{S\subseteq\Omega} \prod_{i = 1}^{N} p(y^i|S^i).
\end{align}

It can be seen that maximizing $I(S;Y)$ is equivalent to maximizing the conditional likelihood of the label $Y$ given the selected feature subset $S$. The conditional likelihood is a well-studied statistical principle and maximizing conditional likelihood is the objective of may statistical models (e.g., Naive Bayes Classifier \cite{friedman1997}). Therefore, selecting the feature subset $S$ that maximizes MI with the label can help to improve the classification performance.

\begin{figure*}[t!]
\renewcommand\arraystretch{1.25}
\centering
\begin{tikzpicture}[baseline=(current bounding box.center)]
\pgfplotsset{width=11.2cm,height=5.4cm}
\pgfplotsset{ticklabel style={/pgf/number format/precision=4}, tick scale binop={\times}}
	\tikzstyle{block} = [rectangle, draw, fill=white!20, 
	text width=10em, text centered, rounded corners, minimum height=1em, minimum width=40mm]
	
	\tikzstyle{empty} = [rectangle, draw = white, fill=white!20, 
	text width=4em, text centered, rounded corners, minimum height=0em, minimum width=0mm]
	
	\tikzstyle{bigbox} = [rectangle, minimum width=1cm, minimum height=3cm, text centered, text width=1cm, draw=black, fill=white!5]
	\tikzstyle{line} = [draw, color=blue, width=10mm, -latex']
	\tikzstyle{arrow} = [ultra thick,->,>=stealth]
	\node [black] at (4.8,-0.6) (c) {Number of Features in the Selected Feature Subset $S$};
	\node [black, rotate=90] at (-1.2, 1.7) (c) {Joint MI $I(S;Y)$};
\begin{axis}[legend pos=south east]
[
    ytick={0.247,0.248, ..., 0.255},
    ymin=0.247,
    ymax=0.255,
/pgf/number format/precision=5
    ]
    \addplot[
        scatter,only marks,scatter src=explicit symbolic,
        scatter/classes={
            a={mark=*,blue},
            b={mark=triangle*,red},
            c={mark=square*,draw=black,fill=black}
        }
    ]
    table[x=x,y=y,meta=label]{
        x    y    label
       1  0.247  a
       2  0.24827 a
       3  0.2489 a
       4 0.2495 a
       5 0.25 a
       6 0.2505 a
       7 0.2509 a
       8 0.2513 a
       9 0.2517 a
       10 0.2522 a
       11 0.2524 a
       12 0.2526 b
       13 0.2527 a
       14 0.25279 a
       15 0.25287 a
       16 0.25296 a
       17 0.253 a
       18 0.25317 b
       19 0.25321 a
       20 0.25329 b
       21 0.25337 a
       22 0.25346 a
       23 0.25354 a
       24 0.25362 a
       25 0.25371 b
       26 0.25379 c
       27 0.25387 c
       28 0.25396 b
       29 0.25404 c
       30 0.25412 a
       31 0.25420 c
       32 0.25429 a
       33 0.25433 b
       34 0.25445 b
       35 0.25445 b
       36 0.25445 b
       37 0.25445 b
       38 0.25445 b
       39 0.25445 b
       40 0.25445 b
    };
    \legend{UR feature subset $S_{{\scriptscriptstyle UR}}$, CR feature subset $S_{{\scriptscriptstyle CR}}$, Redundant feature subset $S_{red}$},
\end{axis}
\end{tikzpicture}%
\renewcommand\arraystretch{1.75}
\hspace{8mm}\begin{tabular}{|c|c|} \hline
algorithm & redundancy rate ($\gamma$) \\ \hline
MIM        & 23.1\% (9/39)        \\
JMI         & 14.3\% (5/35)      \\
JMIM      & 21.1\%  (8/38)      \\ 
mRMR   & 18.9\% (7/37)       \\
GSA       & 11.8\% (4/34)        \\\hline
\end{tabular}
\caption{ (left) Illustration of feature selection using GSA on the Sonar dataset \cite{Dua2019}. (right) Redundancy rates ($\gamma=|S_{red}| / |S_{sat}|$) for various MIBFS algorithms on the Sonar dataset. The numbers in parentheses are $|S_{red}|$ and $|S_{sat}|$, respectively.}
\label{MI-Var}
\vspace{0mm}
\end{figure*}

\subsection{Goal of MIBFS: Why Minimal Feature Subset?}
\label{sec3.2}
Including more terms than necessary is essentially an overfitting issue defined in \cite{hawkins2004} and it may cause two problems: 


\textbf{(i) Efficiency.} In terms of maximizing joint MI $I(S;Y)$, if $K$ features can maximize the joint MI. Including more redundant features will cause extra computation and storage. 

\textbf{(ii) Overfitting to irrelevance.} More importantly, more features may contain more dominant characteristics of other domains \cite{ding2005}. These dominant characteristics of other domains are often known as irrelevance and overfitting to the irrelevance may confuse the classifier, leading to degraded performance \cite{john1994, song2011}.  The following proposition demonstrates that including extra feature will increase the irrelevance presented in the selected feature subset. 

\begin{prop}
Assuming that $H(X_k|S^*) \neq 0$, including an additional feature $X_k$ to the optimal feature subset $S^*$ in \eqref{NewP} will increase the irrelevance presented in the selected feature subset $\{X_k, S^*\}$.
\begin{proof}
Since $S^*$ is the optimal feature subset in \eqref{NewP}, we have
\begin{align}
&I(S^*;Y) = I(S^*,X_k;Y). \label{eq10}
\intertext{By chain rule for MI, \eqref{eq10} can be derived as} 
&H(S^*) - H(S^*|Y) = H(S^*,X_k) - H(S^*,X_k|Y). \nonumber
\intertext{After manipulation on both sides, we have } 
&H(S^*,X_k|Y) - H(S^*|Y) = H(S^*,X_k) - H(S^*) \nonumber \\
& \hspace{38mm} = H(X_k|S^*) \geq 0. \label{proof2_1} 
\end{align}
\label{proof2}
\vspace{-4mm}
\end{proof}
\end{prop}

Referring to our definition of irrelevance in \eqref{irr}, the irrelevance of feature subset $\{X_k, S^*\}$ is $H(S^*,X_k|Y)$ while the irrelevance of feature subset $S^*$ is $H(S^*|Y)$. In \eqref{proof2_1}, we show that, after including feature $X_k$, the irrelevance with respect to the label $Y$ will increase as long as the entropy of feature $X_k$ does not fully overlap with the joint entropy of $S^*$ (i.e., $H(X_k|S^*) > 0$). Only when the entropy of feature $X_k$ fully overlaps with the joint entropy of $S^*$ (i.e., $H(X_k|S^*) = 0$), including feature $X_k$ will not increase the irrelevance. However, it is unlikely to happen in practice.

\subsection{Goal of MIBFS: A Crucial Condition for Optimality}
\label{sec3.3}
Recall the goal of MIBFS in \eqref{NewP} is to find the minimal feature subset with maximum MI with respect to the label \cite{Brown}. Several works \cite{yu2004efficient,john1994} have pointed out that 
UR is a necessary condition for the optimal solution in \eqref{NewP}. This can be simply proved as follows.
\begin{prop}
The optimal feature subset $S^*$ in \eqref{NewP}, which has maximum $I(S;Y)$ with minimum $|S|$, must contain all features with UR.
\begin{proof}
Assume there exists a feature $X_k \in \Omega $ with non-zero UR.  Suppose we have a feature subset $S \subseteq \Omega \backslash X_k$ which has maximum $I(S;Y)$ with minimum $|S|$. Since $X_k$ has non-zero UR, we have $I(\Omega;Y) > I(\Omega \backslash X_k;Y)$ by definition. Therefore, $I(\Omega;Y) >  I(S;Y)$ as $I(\Omega \backslash X_k;Y) \geqslant  I(S;Y)$ given $S \subseteq \Omega \backslash X_k$. But this contradicts the initial assumption that $I(S;Y)$ is maximum.
\label{proof}
\end{proof}
\end{prop}
We note that the optimal feature subset $S^*$ may also contain features with zero UR at certain situations. For example, consider a feature subset $T$ which contains all features with UR, it is possible to have another feature $X_m \not\in T$ which contributes to higher joint MI (i.e., $I(X_m,T;Y) > I(T;Y)$). The reason for $X_m \not\in T$ could be that $X_m$'s relevance with respect to $Y$ overlaps with other features.

\begin{table*}[t!]
\renewcommand\arraystretch{1.7}
\setlength\tabcolsep{4pt} 
\centering
\normalsize
  \begin{tabular}{|c|c|} \hline \hline
\multicolumn{1}{|c|}{\bf MRwMR based algorithms} & \multicolumn{1}{c|}{\bf MRwMR-BUR based algorithms} \\ \hline 
$J_{\text{MIM}} (X_i) =  I(X_{i};Y)$ & $J_{\text{MIM-BUR}}(X_i) = (1-\beta) \times I(X_{i};Y) + \beta \times J_{\text{UR}}(X_i)$ \\ 
$J_{\text{JMI}}(X_i) = \underset{X_j \in S}{\sum}I(X_i, X_j; Y)$ & $J_{\text{JMI-BUR}}(X_i) = (1-\beta) \times \underset{X_j \in S}{\sum}I(X_i, X_j; Y) \times \frac{1}{|S|} + \beta \times J_{\text{UR}}(X_i)$ \\

$ J_{\text{mRMR}}(X_i) = I(X_i;Y) - \frac{1}{|S|} \underset{X_j \in S}{\sum} I(X_i, X_j) $ & $J_{\text{mRMR-BUR}}(X_i) = (1-\beta) \times \big(I(X_i;Y) - \frac{1}{|S|} \underset{X_j \in S}{\sum} I(X_i, X_j)\big) + \beta \times J_{\text{UR}}(X_i)$ \\
$ J_{\text{JMIM}}(X_i) = \underset{X_j \in S}{\min} I(X_i, X_j;Y) $ & $ J_{\text{JMIM-BUR}}(X_i) = (1-\beta) \times (\underset{X_j \in S}{\min} I(X_i, X_j;Y)) + \beta \times J_{\text{UR}}(X_i) $ \\
$ J_{\text{GSA}}(X_i) =  I(X_i,S;Y) $ & $ J_{\text{GSA-BUR}}(X_i) = (1- \beta) \times I(X_i,S;Y) +  \beta \times J_{\text{UR}}(X_i) \label{GSA-BUR}$ \\ \hline \hline

  \end{tabular}
  \captionsetup{justification=raggedright,skip=2pt}
  \caption{The scoring functions of various MRwMR based algorithms and their corresponding MRwMR-BUR forms. Depending on the task of MRwMR-BUR, UR can be estimated differently, leading to two variants of MRwMR-BUR: (i) MRwMR-BUR-KSG; (ii) MRwMR-BUR-CLF.}
  \label{all_algorithms}
\end{table*}

\subsection{Evaluation of MRwMR based Algorithms}
\label{sec3.4}
In this subsection, we evaluate the performance of five representative MIBFS algorithms (described in Section \ref{sec2.3}) in achieving the goal of MIBFS described in \eqref{NewP}. Specifically, we aim to count how many redundant features are selected by each of them when the joint MI $I(S;Y)$ firstly saturates. We simulate their feature selection process using the Sonar dataset \cite{Dua2019} and evaluate the variation of joint MI $I(S;Y)$ as more features are selected. Let $S_{sat}$ be the selected feature subset when $I(S;Y)$ firstly reaches saturation. We further divide $S_{sat}$ into two subsets as follows: $S_{sat} = \{ S_{{\scriptscriptstyle UR}}, S_{{\scriptscriptstyle ZUR}} \}$.
$S_{{\scriptscriptstyle UR}}$ is the UR feature subset and contains selected features with non-zero UR.
$S_{{\scriptscriptstyle ZUR}}$ is the zero UR feature subset contains selected features with zero UR.


Since $S_{{\scriptscriptstyle UR}}$ is necessarily a subset of the optimal subset $S^*$ in \eqref{NewP},  we only conduct exhaustive search on $S_{{\scriptscriptstyle ZUR}}$ and find the largest feature subset $S_{red}$ which can be removed without decreasing the joint MI. In such a manner,  we evaluate the feature subset selected by various MIBFS algorithms. Mathematically, the problem is formulated as follows.
\begin{equation} 
S_{{red}}= \mathop{\arg \max} \\ f(\mathop{\arg\max}_{\bar{S}\subseteq S_{{\scriptscriptstyle ZUR}}} I (S_{{\scriptscriptstyle UR}}, S_{{\scriptscriptstyle ZUR}} \backslash \bar{S} ;Y)),
\end{equation}
where $f (A,B,\cdots) = (|A|,|B|,\cdots)$, $|A|$ represents the number of features in $A$. By obtaining $S_{red}$, we further divide $S_{{\scriptscriptstyle ZUR}}$ into two subsets  (i.e., $S_{{\scriptscriptstyle ZUR}} = \{ S_{{\scriptscriptstyle CR}}, S_{red} \}$):  

\begin{enumerate}[noitemsep,leftmargin=5mm, topsep=0pt]
\item The redundant feature subset $S_{red}$: the maximal feature subset which can be removed from $S_{{\scriptscriptstyle ZUR}}$ without decreasing the joint MI (i.e., $I(S_{{\scriptscriptstyle UR}}, S_{{\scriptscriptstyle ZUR}} ; Y) = I(S_{{\scriptscriptstyle UR}}, S_{{\scriptscriptstyle ZUR}} \backslash S_{red} ; Y$)).

\item The CR feature subset $S_{{\scriptscriptstyle CR}}$: the minimal feature subset which provides the maximum MI (joint with $S_{{\scriptscriptstyle UR}}$) with the label $Y$. The feature subset $S_{{\scriptscriptstyle CR}}$ contains selected features which have CR with the label $Y$ given $S_{{\scriptscriptstyle UR}}$.
\end{enumerate}


\noindent To quantify the redundancy of a selected feature subset, we introduce a term called redundancy rate ($\gamma$), represented as
\begin{equation}
\gamma = |S_{red}| / |S_{sat}|.
\end{equation}

In Fig. \ref{MI-Var} (left), we evaluate the variation of joint MI $I(S;Y)$ as more features are selected by GSA. The performance of all selected MRwMR based algorithms summarized using redundancy rate is shown in Fig. \ref{MI-Var} (right). A higher redundancy rate is undesirable as it indicates that more redundant features are selected.


In Fig. \ref{MI-Var} (left), we observe that GSA can perform well at the beginning and select features with UR. However, as more features are selected, the joint distribution of the selected feature subset becomes more complex, and then GSA tends to select redundant features. Similarly, a non-negligible number of redundant features are also selected by other algorithms (see Fig. \ref{MI-Var} (right)). We note that redundant features contribute nothing, but they do increase the size of the selected feature subset, undermining the objective of minimal feature subset in \eqref{NewP}. Surprisingly, all of the representative algorithms studied select a non-negligible number of redundant features, which uncovers the fact that all of them are underperforming. 
Lastly, we note that we are able to find $S_{red}$ on $S_{{\scriptscriptstyle ZUR}}$ via an exhaustive search because the Sonar dataset is a relatively low dimensional dataset. In general, most real world datasets tend to contain more features and it is computationally infeasible to conduct such an exhaustive search.

\begin{table*}[t!]
\normalsize
\renewcommand\arraystretch{1.5}
\setlength\tabcolsep{14pt} 
\centering
  \begin{tabular}{l|ccccc} \hline
  		\hline             
		& $\beta$ = 0 & $\beta$ = 0.1 & $\beta$ = 0.2 & $\beta$ = 0.5 & $\beta$ = 0.9 \\ \hline
		MIM-BUR-KSG        & 23.1\% (9/39) & {\bf 18.9\%} (7/37) & 21.1\% (8/38) & 23.1\% (9/39) & 23.1\% (9/39) \\ 
		JMI-BUR-KSG         & 14.3\% (5/35) & 16.7\% (6/36) &  {\bf 11.8\%} (4/34) & 18.9\% (7/37) & 21.1\% (8/38) \\ 
		JMIM-BUR-KSG      & 21.1\% (8/38)  & {\bf 16.7\%} (6/36) & 18.9\% (7/37) & 23.1\% (9/39) & 23.1\% (9/39) \\ 
		mRMR-BUR-KSG   &  18.9\% (7/37)   & {\bf 14.3\%} (5/35) & 16.7\% (6/36) & 23.1\% (9/39) & 25\% (10/40) \\
		GSA-BUR-KSG       & 11.8\% (4/34)   & {\bf 6.25\%} (2/32) & 14.3\% (5/35) & 16.7\% (6/36) & 16.7\% (6/36) \\ 	\hline
		\cmidrule{1-6}
  \end{tabular}
  \captionsetup{justification=raggedright,skip=0pt}
  \caption{Redundancy Rate ($\gamma=|S_{red}|$ / $|S_{sat}|$) of MRwMR based Algorithms before ($\beta$ = 0) and after Boosting UR (see \eqref{MRwMR-BUR}) on the Sonar dataset \cite{Dua2019}. All the MI quantities are estimated using the KSG estimator. The numbers in parentheses are $|S_{red}|$ and $|S_{sat}|$, respectively. The numbers in bold represent the lowest redundancy rate over different values of $\beta$.}
  \label{redundancy_exp_new}
\end{table*}

\begin{table*}[!b]
\normalsize
\setlength\tabcolsep{10.4pt}
{\renewcommand{\arraystretch}{1.2}
\centering
\begin{tabular}{l|cccccccc} \hline
\hline                       
 & Colon \cite{alon1999broad} & Sonar \cite{Dua2019} & Madelon \cite{madelon} & Leukemia \cite{golub1999molecular} & Isolet  \cite{Dua2019}  & Gas sensor \cite{Alex2012}\\ \hline
Features   & 2000   & 60 &  500 & 7070 & 617 & 128 \\ 
Instances  & 62      & 208 &  2600 & 72 & 1560 & 13874\\ 
Classes     & 2        & 2 & 2  & 2 & 26  & 6\\ 
Data Type & Discrete  & Continuous & Continuous & Discrete & Continuous & Continuous\\ \hline
UR (\%)     & 4.3\% & 28.3\% &29.6\% & 34.9\% & 37.1\% & 2.34\% \\ \hline \hline
\end{tabular}
\vspace{-2mm}
\caption{Information of the six public datasets used in the experiments. The last row computes the percentage of features with UR among all features for each dataset.}
\label{data}
}
\end{table*}

\subsection{The Proposed MRwMR-BUR Criterion}
\label{sec3.5}

{\bf (1) Motivation for MRwMR-BUR. }We point out that the heart of the problem is that features with UR are not prioritized during the selection process as all these algorithms follow the MRwMR criterion which lacks a mechanism to identify UR. Furthermore, based on our analysis of various datasets (see Table \ref{data} row 5), features with UR usually make up a very small fraction of the total number of features and hence, is difficult to select them without explicitly targeting them. This motivates us to augment MRwMR and include the objective of boosting unique relevance (BUR), leading to a new criterion, called MRwMR-BUR. 

{\bf (2) The MRwMR-BUR Criterion. }The augmented MRwMR-BUR based algorithms are
\begin{align}
\label{MRwMR-BUR}
J_{new}(X_i) = (1-\beta)\times J_{org}(X_i) + \beta \times J_{{\scriptscriptstyle UR}}(X_i),
\end{align}
where $J_{org}$ is the original MRwMR based algorithm (e.g., MIM), $J_{new}$ is corresponding MRwMR-BUR form and $J_{{\scriptscriptstyle UR}}(X_i)$ returns the UR of feature $X_i$. Depending on the task being addressed, MRwMR-BUR has two variants, termed {\bf MRwMR-BUR-KSG} and {\bf MRwMR-BUR-CLF}, which estimate UR differently. MRwMR-BUR-KSG directly estimates UR using \eqref{URE} via a nearest neighbor based approach called KSG estimator and is designed for three major tasks: (i) Classification Performance (i.e., higher classification accuracy). (ii) Feature Interpretability (i.e., a more precise feature subset to explore the hidden relationship between features and labels). (iii) Classifier Generalization (i.e., the selected feature subset generalizes well to various classifiers). MRwMR-BUR-CLF estimates UR via a classifier based approach, aiming the further improve the competitiveness of MRwMR-BUR for classification performance oriented tasks. More details of MRwMR-BUR-CLF are discussed in Section \ref{sec5}.

We note that modification of MRwMR based algorithms in \eqref{MRwMR-BUR} is slightly different for JMI as JMI is not bounded and the score increases as more features are selected. Therefore, we divide the original form of JMI by the size of the selected feature subset and include BUR. Moreover, the details for each algorithm with and without BUR are provided in Table \ref{all_algorithms}. We will denote the algorithm that extends XYZ as XYZ-BUR (e.g., MIM and MIM-BUR). Furthermore, if the UR is estimated via the KSG estimator, we further extend XYZ-BUR as XYZ-BUR-KSG. Likewise, XYZ-BUR-CLF indicates that the UR is estimated via the classifier based approach.

{\bf (3) MRwMR-BUR reduces the redundancy rate. }In Table \ref{redundancy_exp_new}, we estimate all the MI quantities using the KSG estimator and present the redundancy rate of five representative MRwMR based algorithms before and after boosting UR using the same Sonar dataset in Section \ref{sec3.4}. We note that the MRwMR-BUR based algorithm degenerates to the original MRwMR form when $\beta$ = 0 (see \eqref{MRwMR-BUR}). It can be seen that the redundancy rate is reduced after slightly boosting UR (e.g., $\beta$ = 0.1). For example, the number of redundant features selected by GSA is reduced from 4 to 2 after boosting features with UR ($\beta = 0.1$), leading to a lower redundancy rate of 6.25\%. On the other hand, when we heavily boost features with UR by increasing the value of $\beta$ to 0.5 and 0.9, it causes an increase in the redundancy rate. We posit it is because that heavily boosting features with UR could severely destroy the mechanism of the original MRwMR based algorithm. Specifically, heavily boosting features with UR will only select the candidate feature with UR and overlook its CR with respect to the label Y given $S$. 


{\bf (4) Complexity of MRwMR-BUR. }In terms of the computational complexity, MRwMR-BUR based algorithms have the same complexity as the corresponding MRwMR based algorithms. This is because the value of UR only needs to be calculated once and can be used in subsequent calculations. 

\section{Performance Evaluation}
\label{sec4}
In Section \ref{sec3.3}, we motivated the idea of MRwMR-BUR and demonstrated that MRwMR-BUR could significantly reduce the redundancy rate. In this section, we estimate UR using the KSG estimator and compare the performance of each algorithm before and after boosting UR, so as to evaluate the MRwMR-BUR criterion. 
In Section \ref{sec4.1}, we describe the experiment setup. Next, in Section \ref{sec4.3}, we conduct the performance comparison and analyze the results.

\subsection{Experiment Setup}
\label{sec4.1}
{\bf (1) Experiment Details. }To examine the performance of the proposed MRwMR-BUR criterion, we conduct experiments using six public datasets \cite{alon1999broad,golub1999molecular,Dua2019,madelon,Alex2012} (see descriptions in Table \ref{data}) and compare the performance of MRwMR-BUR-KSG (estimate UR via the KSG estimator) to MRwMR via three popular classifiers: Support Vector Machine (SVM) \cite{SVM}, K-Nearest Neighbors (KNN) \cite{KNN} and Random Forest (RF) \cite{RF}. Five representative MRwMR based algorithms: MIM \cite{Lewis1992}, JMI \cite{Yang1999}, JMIM \cite{Bennasar2015}, mRMR \cite{Peng2005} and GSA\cite{Brown} are shortlisted for performance evaluation. 

For each run, the dataset is randomly split into three subsets: training dataset (60\%), validation dataset (20\%), testing dataset (20\%). We apply MRwMR and MRwMR-BUR-KSG (with $\beta$ = 0.1) based algorithms on the same training dataset to select features and evaluate them using the same testing dataset. We compute the validation accuracy for gradually selecting up to $k$ features, resulting in a vector of $[\theta_1, \theta_2, \cdots, \theta_k]$.  Next, we shortlist the first $n$ features which provide the highest validation accuracy and use them to compute the test accuracy. The test accuracy averaged over 20 runs are shown for RF (in rows 1 - 10), KNN (in rows 11 - 20), and SVM (in rows 21 - 30). Furthermore, the round-up average number of features chosen (i.e., the value in parentheses) is also an indication of the performance in terms of the objective of minimal feature subset in \eqref{NewP}.

\begin{table*}[!t]
\normalsize
\renewcommand\arraystretch{1.42}
\setlength\tabcolsep{32pt} 
\centering
  \begin{tabular}{lccc} \hline
  		\hline             
		& KNN & SVM & RF \\ \hline
		MM-BUR-KSG \{$S_1$\}                                                             & 85.0\%   & 82.2\% & 84.8\%\\ 	
		MIM-BUR-KSG \{$S_1$ $\backslash$ \{Hsa.8, Hsa.1132\}\}      & 83.2\%   & 80.7\% & 84.1\% \\  \hline	 	
		GSA-BUR-KSG \{$S_2$\}                                                            & 85.5\%   & 74.5\% & 86.3\% \\ 	
		GSA-BUR-KSG \{$S_2$ $\backslash$ \{Hsa.8, Hsa.1132\}\}      & 84.1\%   & 73.8\% & 85.0\% \\  \hline \hline
  \end{tabular}
  \captionsetup{justification=raggedright,skip=0pt}
  \caption{Test Accuracy of MIM-BUR-KSG and GSA-BUR-KSG with and without two important features (Hsa.8 and Hsa.1132) on the Colon dataset via KNN, SVM and RF. $S_1$ and $S_2$ represent the feature subset with maximum validation accuracy selected by MIM-BUR-KSG and GSA-BUR-KSG, respectively. Both Hsa.8 and Hsa.1132 have a relatively lower MI, which makes them hard to be selected by MRwMR based algorithms. With MRwMR-BUR, these two features are selected, resulting in a higher accuracy and a more precise feature subset to explore hidden relationships between features and labels.  }
  \label{accuracy_w_wo}
\end{table*}

{\bf (2) Parameter Tuning. }To ensure fair comparison, the parameters of all classifiers are tuned using the validation dataset via grid search and all algorithms share the same grid searching range and step size. Some key parameters are tuned as follows. (i) the number of nearest neighbors $K$ for KNN is tuned from 3 to 50 with step size of 2. (ii) the regularization coefficient $c$ for SVM is chosen from \{0.001, 0.01, 0.1, 1, 10\}. (iii) the number of decision trees in the RF is chosen from \{10, 15, ..., 100\}. We directly estimate the UR of features using \eqref{URE} via the KSG estimator. This is the same as how we estimate UR in Sections \ref{sec3.2} and \ref{sec3.3}. Following the same notation, we denote the algorithm that extends XYZ as XYZ-BUR-KSG (e.g., MIM and MIM-BUR-KSG). 

{\bf (3) Source Code. }We highlight that the source code will be released to allow reproducibility by the research community.

\subsection{Performance Comparison}
\label{sec4.3}
In Table \ref{tab:rf}, we compare the performance between MRwMR based algorithms and their corresponding MRwMR-BUR-KSG algorithms. The results are presented as average test accuracy $\pm$ standard deviation over 20 runs and the number in the parentheses represents the number of features chosen. 

\textbf{(1) Performance of MRwMR based Algorithms.} In terms of MRwMR based algorithms, we observe that MIM tends to provide the worst performance in terms of average test accuracy and number of features required (e.g., see Colon dataset). We suspect it is because that MIM assumes features are independent from each other, leading to degraded performance. For other MRwMR based algorithms, JMI and GSA generally perform better than other MRwMR based algorithms (e.g., see Madelon dataset). This finding agrees with \cite{Brown, shiyuBIBM} as GSA is greedy in nature and JMI can increase the complementary information between features. 

\textbf{(2) Performance of MRwMR VS MRwMR-BUR-KSG.} Comparing the performance of MRwMR to MRwMR-BUR-KSG, we observe that most of the MRwMR based algorithms improve their performance after BUR. As an example, the accuracy of GSA on Colon dataset using RF  is increased from 84.7\% to 86.5\% (compare row 9 to row 10 in Table \ref{tab:rf}), resulting an improvement of 2.1\%. In addition, for the dataset that we have been able to obtain extremely high accuracy (i.e., the Gas sensor dataset with 99\%+ accuracy), MRwMR-BUR can significantly reduce the number of features required while maintaining comparable performance. As an example, the number of features required for mRMR on the Gas sensor dataset using RF (compare row 3 to row 4 in Table \ref{tab:rf}) is decreased from 93 to 69, a reduction of 25.8\%.

\textbf{(3) Generalization to Other Classifiers.} In addition to the performance using RF mentioned above, the superior performance of MRwMR-BUR-KSG tends to generalize to KNN and SVM as well. As an example, the accuracy of mRMR on the Colon dataset using KNN is increased from 82.3\% to 84.7\% after BUR (compare row 13 to row 14), resulting in an improvement of 3\%. The number of features needed is decreased from 43 to 30, a reduction of 30.2\%.

\textbf{(4) Feature Interpretability. } As compared to MRwMR, MRwMR-BUR-KSG helps to better explore hidden relationships between features and labels in two aspects: 

(i) The feature subset selected by MRwMR-BUR-KSG based algorithms contains fewer number of features and excludes more noise from redundant or irrelevant features. This can be seen from above where MRwMR-BUR-KSG based algorithms select a much smaller number of features. 

(ii) MRwMR-BUR slightly prioritizes features with UR, leading to a more precise feature subset for study. As a detailed example, MRwMR-BUR-KSG helps to identify two important features in the Colon dataset with 2000 gene features: Hsa.8 (Human mRNA for ORF) and Hsa.1132 (Human mRNA for hepatoma-derived growth factor). Both of these two features have a relatively lower MI, which cause them not to be selected by any of the five representative algorithms studied when those algorithms achieve the peak accuracy. However, both of them are features with UR, which can be identified and selected by MRwMR-BUR-KSG algorithms. This leads to a higher test accuracy on all classifiers studied (i.e., in Table \ref{accuracy_w_wo}, compare rows 1 and 3 to rows 2 and 4,  respectively), suggesting the crucial role of these two features in diagnosis (tumoral and non-tumoral). We have discussed our findings with domain experts. The feedback is that such findings are useful which could provide medical staff with some new insights and further advance relevant research. In the table \ref{table_5} below, we summarize similar features, i.e., features with UR that can lead to better diagnosis, but are often ignored by MRwMR based algorithms due to their relatively low MI.
\begin{table}[!b]
\vspace{-2mm}
\normalsize
\renewcommand\arraystretch{1.4}
\setlength\tabcolsep{14pt} 
\centering
\begin{tabular}{l|cc} \hline
  		\hline             
		Dataset & Feature (Description) \\ \hline
Leukemia & U22376 (C-myb), M31523 (E2A),  \\ 
                &  M69043 (MAD-3), U46751 (p62). \\ \hline	
Colon & Hsa.3307 (Human Gps2 mRNA) \\
          & Hsa.2598 (H.sapiens B-cam mRNA) \\  \hline \hline	 	
  \end{tabular}
  \captionsetup{justification=raggedright,skip=0pt}
  \caption{A List of features identified by MRwMR-BUR-KSG which could contribute to better feature interpretability.}
  \label{table_5}
\end{table}

\textbf{(5) Algorithm Complexity.} We highlight that the value of UR only needs to be calculated once and can be used in subsequent calculations. Hence, the algorithm complexity after BUR is comparable to the original algorithm.


\begin{table*}[!t]
\setlength\tabcolsep{5.5pt}
{\renewcommand{\arraystretch}{2.0}
\centering
\begin{tabular}{l|cccccccc} \hline
\hline      
 & Colon & Sonar & Madelon & Leukemia & Isolet   & Gas sensor\\ \hline  
\multicolumn{7}{c}{\bf \em Random Forest (RF)} \\ \hline
1) MIM                      & 83.7 $\pm$ 7.3 (48)    & 79.5 $\pm$ 3.4 (43)   & 71.5 $\pm$ 1.3 (86)  & 95.5 $\pm$ 4.3 (84) & 85.1 $\pm$ 0.8 (142)     & 99.27 $\pm$ 0.13 (86) \\
2) MIM-BUR-KSG             & {\small \textbf{85.1 $\pm$ 6.1 (23)}}    & {\small \textbf{80.1 $\pm$ 2.8 (39)}}   & {\small \textbf{72.4 $\pm$ 1.5 (76)}}  & {\small \textbf{96.5 $\pm$ 2.3 (45)}}  & {\small \textbf{86.1 $\pm$ 1.5 (134)}}    & {\small \textbf{99.43 $\pm$ 0.05 (85)}} \\ \hline

3) mRMR                    &  85.0 $\pm$ 5.1 (14)  & 79.4 $\pm$ 2.5 (46)   & 72.0 $\pm$ 0.8 (76) & 96.1 $\pm$ 3.3 (76)  & 85.8 $\pm$ 0.3 (121)   & 99.44 $\pm$ 0.04 (93) \\
4) mRMR-BUR-KSG         &  {\small \textbf{87.0 $\pm$ 3.1 (8)}}    & {\small \textbf{81.0 $\pm$ 2.0 (39)}}   & {\small \textbf{72.4 $\pm$ 0.3 (74)}}  & {\small \textbf{96.5 $\pm$ 3.1 (61)}}  & {\small \textbf{86.3 $\pm$ 0.5 (125)}}   & {\small \textbf{99.46 $\pm$ 0.03 (69)}} \\ \hline

5) JMI                       & 72.8 $\pm$ 2.1 (31)   & 80.5 $\pm$ 3.1 (42)   & {\small \textbf{72.5 $\pm$ 0.3 (60)}}  & 95.6 $\pm$ 1.3 (59) &  87.0 $\pm$ 0.2 (130)     & 99.45 $\pm$ 0.06 (79) \\
6) JMI-BUR-KSG              & {\small \textbf{74.7 $\pm$ 6.0 (37)}}   & {\small \textbf{81.0 $\pm$ 3.9 (46)}}   & 71.7 $\pm$ 0.3 (61)  & {\small \textbf{97.1 $\pm$ 1.4 (63)}}  & {\small \textbf{87.8 $\pm$ 0.5 (133)}}    & {\small {\bf 99.46 $\pm$ 0.04 (65)}}  \\ \hline

7) JMIM                    & 73.2 $\pm$ 4.7 (33)  & 79.6 $\pm$ 1.3 (46)   & 72.8 $\pm$ 0.5 (70)   & 95.6 $\pm$ 3.3 (58)  & 85.3 $\pm$ 0.9 (123)    & 99.44 $\pm$ 0.03 (86)     \\
8) JMIM-BUR-KSG           & {\small \textbf{75.7 $\pm$ 2.9 (20)}}  & {\small \textbf{80.6 $\pm$ 1.4 (36)}}   & {\small \textbf{73.1 $\pm$ 0.4 (59)}}  & {\small \textbf{95.9 $\pm$ 1.0 (59)}}  & {\small \textbf{85.7 $\pm$ 0.3 (119)}}    & {\small \textbf{99.48 $\pm$ 0.09 (75)}}   \\ \hline

9) GSA                     & 84.7 $\pm$ 2.3 (19)  & 80.3 $\pm$ 3.1 (25)  & 73.0 $\pm$ 0.6 (64)   & {\small \textbf{95.6 $\pm$ 3.1 (60)}} & 86.2 $\pm$ 0.3 (118)    &99.42 $\pm$ 0.04 (95)  \\
10) GSA-BUR-KSG            & {\small \textbf{86.5 $\pm$ 1.5 (14)}}  &  {\small \textbf{81.7 $\pm$ 2.9 (27)}}  & {\small {\bf 73.4 $\pm$ 0.4 (57)}}   & 95.5 $\pm$ 1.3 (55)  & {\small \textbf{87.2 $\pm$ 0.3 (115)}}   & {\small \textbf{99.47 $\pm$ 0.04 (85)}}    \\ \hline

 \multicolumn{7}{c}{\bf \em K-Nearest Neighbors (KNN)} \\ \hline

 11) {MIM}                             & 84.0 $\pm$ 5.1 (60)                & 82.6 $\pm$ 1.3 (24)                  & 73.0 $\pm$ 0.8 (55)              & 95.3 $\pm$ 1.2 (44)                & 78.1 $\pm$ 0.9 (117)            & 99.04 $\pm$ 0.03 (85)  \\ 
 12) {MIM-BUR-KSG}                    & {\small \textbf{85.2 $\pm$ 3.9 (40)}}                & {\small \textbf{83.2 $\pm$ 1.1 (24)}}                  & {\small \textbf{74.5 $\pm$ 1.2 (49)}}              & {\small \textbf{96.0 $\pm$ 1.5 (38)}}                & {\small \textbf{78.7 $\pm$ 0.6 (129)}}            &  {\small \textbf{99.09 $\pm$ 0.06 (83)}} \\  \hline

 13) {mRMR}                        & 82.3 $\pm$ 3.7 (43)                & 84.8 $\pm$ 2.0 (33)                  & 75.3 $\pm$ 1.5 (59)               & 97.5 $\pm$ 0.7 (56)                & 80.8 $\pm$ 1.3 (123)             & 98.94 $\pm$ 0.02 (80) \\
 14) {mRMR-BUR-KSG}               & {\small \textbf{84.7 $\pm$ 2.9 (30)}}                & {\small \textbf{85.6 $\pm$ 1.3 (33)}}     & {\small \textbf{76.0 $\pm$ 2.0 (65)}}               & {\small \textbf{97.6 $\pm$ 0.8 (52)}}                & {\small \textbf{81.9 $\pm$ 1.6 (109)}}              & {\small \textbf{99.02 $\pm$ 0.09 (78)}}  \\ \hline

 15) {JMI}                             & 70.0 $\pm$ 2.5 (45)                  & 83.2 $\pm$ 1.7 (45)                & 79.0 $\pm$ 1.5 (49)               & 95.4 $\pm$ 1.1 (44)                & {\small \textbf{79.6 $\pm$ 1.1 (127)}}             & 99.08 $\pm$ 0.06 (70) \\
 16) {JMI-BUR-KSG}                    & {\small \textbf{70.4 $\pm$ 2.7 (40)}}                  & {\small \textbf{84.8 $\pm$1.3 (42)}}    & {\small \textbf{79.8 $\pm$ 1.2 (51)}}    & {\small \textbf{96.3 $\pm$ 0.7 (44)}}                & 79.3 $\pm$ 1.0 (127)             & {\small \textbf{99.10 $\pm$ 0.05 (63)}}  \\ \hline

 17) {JMIM}                          & 72.0 $\pm$ 2.3 (81)                & {\small \textbf{83.9 $\pm$ 2.1 (43)}}                 & 77.1 $\pm$ 0.6 (55)               & 95.6 $\pm$ 0.5 (32)                & 79.0 $\pm$ 0.7 (135)              & {\small \textbf{99.08 $\pm$ 0.06 (83)}}  \\
 18) {JMIM-BUR-KSG}                 & {\small \textbf{75.9 $\pm$ 2.7 (62)}}                 & 83.7 $\pm$ 1.3 (42)                 & {\small \textbf{78.2 $\pm$ 1.1 (64)}}               & {\small \textbf{95.7 $\pm$ 0.9 (33)}}                & {\small \textbf{79.3 $\pm$ 1.2 (135)}}              & 99.05 $\pm$ 0.03 (78) \\  \hline

 19) {GSA}                           & 84.8 $\pm$ 3.1 (70)                 & 83.3 $\pm$ 1.7 (46)                & 77.5 $\pm$ 1.4 (51)                & 95.4 $\pm$ 1.2 (55)                & 79.7 $\pm$ 1.5 (130)              & 98.7 $\pm$ 0.06 (53) \\ 
 20) {GSA-BUR-KSG}                  & {\small \textbf{86.1 $\pm$ 2.4 (55)}}                  & {\small \textbf{84.4 $\pm$ 1.3 (38)}}               & {\small \textbf{78.3 $\pm$ 1.5 (43)}}                & {\small \textbf{95.8 $\pm$ 0.7 (54)}}                & {\small \textbf{80.1 $\pm$ 1.1 (107)}}               & {\small \textbf{99.0 $\pm$ 0.04 (53)}} \\ \hline

 \multicolumn{7}{c}{\bf \em Support Vector Machine (SVM)} \\ \hline
 
 21) {MIM}                           & 81.0 $\pm$ 3.9 (65)   & 72.8 $\pm$ 1.3 (34)  & 61.2 $\pm$ 0.7 (33)  & 96.4 $\pm$ 1.1 (97)  & 88.0 $\pm$ 0.7 (138)  & 96.5 $\pm$ 0.07 (91)    \\ 
 22) {MIM-BUR-KSG}                  & {\small \textbf{82.0 $\pm$ 2.4 (61)}}   & {\small \textbf{73.6 $\pm$ 1.5 (35)}}  & {\small \textbf{61.6 $\pm$ 0.5 (34)}}  & {\small \textbf{97.6 $\pm$ 1.5 (94)}}  & {\small \textbf{88.6 $\pm$ 1.0 (117)}}  & {\small \textbf{96.6 $\pm$ 0.04 (91)}}   \\ \hline

 23) {mRMR}                       & {\small \textbf{83.0 $\pm$ 2.5 (33)}}   & 75.0 $\pm$ 1.8 (44)  & 61.5 $\pm$ 0.9 (37)  & 97.5 $\pm$ 1.1 (73)  & 89.2 $\pm$ 1.1 (123)  & 96.5 $\pm$ 0.05 (95)    \\
 24) {mRMR-BUR-KSG}              & 82.8 $\pm$ 2.1 (40)   & {\small \textbf{75.3 $\pm$ 1.3 (40)}}  & {\small \textbf{61.5 $\pm$ 0.9 (37)}}  & {\small \textbf{97.5 $\pm$ 1.1 (73)}}  & {\small \textbf{89.6 $\pm$ 1.3 (127)}}  & {\small \textbf{96.8 $\pm$ 0.06 (95)}}  \\ \hline

 25) {JMI}                            & 73.1 $\pm$ 2.2 (15)   & 74.0 $\pm$ 1.1 (39)  & 61.0 $\pm$ 1.3 (33)  & 96.5 $\pm$ 1.4 (83)   & {\small \textbf{89.8 $\pm$ 1.8 (127)}} & 96.7 $\pm$ 0.06 (88)    \\
 26) {JMI-BUR-KSG}                   & {\small \textbf{73.9 $\pm$ 2.9 (15)}}   & {\small \textbf{74.2 $\pm$ 1.6 (37)}}  & {\small \textbf{62.0 $\pm$ 1.2 (35)}}  & {\small \textbf{96.8 $\pm$ 1.4 (86)}}  & 89.4 $\pm$ 1.4 (144)  & {\small \textbf{96.8 $\pm$ 0.09 (89)}}   \\ \hline

 27) {JMIM}                         & 76.6 $\pm$ 2.9 (20)  & 73.1 $\pm$ 1.9 (40)  & 62.0 $\pm$ 0.7 (40)  & 96.7 $\pm$ 0.8 (92)  & 88.3 $\pm$ 1.6 (135)   & 96.5 $\pm$ 0.09 (95)    \\
 28) {JMIM-BUR-KSG}                & {\small \textbf{77.5 $\pm$ 2.1 (24)}}  & {\small \textbf{73.1 $\pm$ 2.4 (36)}}  & {\small \textbf{62.3 $\pm$ 0.9 (43)}}  & {\small \textbf{97.1 $\pm$ 0.8 (93)}}  & {\small \textbf{88.6 $\pm$ 1.7 (115)}}   & {\small \textbf{96.6 $\pm$ 0.09 (96)}}    \\ \hline

 29) {GSA}                          & 73.6 $\pm$ 2.4 (33)  & 73.7 $\pm$ 1.4 (35)  & {\small \textbf{63.3 $\pm$ 1.2 (37)}}  & 96.4 $\pm$ 1.1 (65)  & 89.1 $\pm$ 1.3 (132)   & 96.2 $\pm$ 0.04 (93)    \\ 
 30) {GSA-BUR-KSG}                 & {\small \textbf{74.1 $\pm$ 2.7 (28)}}  &  {\small \textbf{74.0 $\pm$ 1.8 (30)}}  & 63.1 $\pm$ 0.8 (32)  & {\small \textbf{97.0 $\pm$ 1.5 (53)}}  & {\small \textbf{89.5 $\pm$ 1.7 (125)}}   & {\small \textbf{96.4 $\pm$ 0.05 (91)}}    \\  \hline \hline

\end{tabular}
\captionsetup{justification=raggedright,skip=5pt}
\vspace{0mm}
\caption{Performance comparison between MRwMR based algorithms and MRwMR-BUR-KSG based algorithms (with $\beta$ = 0.1) on six public datasets. The results are presented as follows: average test accuracy (\%) $\pm$ standard deviation over 20 runs (\%) (number of features required). The results are shown for RF (row 1 - row 10), KNN (row 11 - row 20) and SVM (row 21 - row 30). All the MI quantities (including UR) are estimated using the KSG estimator. The bold indicates higher performance between MRwMR and MRwMR-BUR-KSG.} 
\label{tab:rf}
}
\end{table*}

\section{Adapting UR to Different Classifiers}
\label{sec5}
In Section \ref{sec4}, we have shown the superior performance of MRwMR-BUR-KSG in accomplishing three major tasks: (i) Classification Performance (ii) Feature Interpretability (iii) Classifier Generalization. In this section, we propose a classifier based approach to estimate UR. This approach of estimating UR adapts UR to different classifier, further improving the competitiveness of MRwMR-BUR for classification performance oriented tasks. 

In Section \ref{sec5.1}, we introduce the classifier based approach to estimate UR which could adapt UR to different classifiers. Next, in Sections \ref{sec5.2} and \ref{sec5.3}, we conduct experiments and compare the classification performance of the classifier based approach to estimating UR to that of using the KSG estimator and three popular classifier based feature selection method.

\subsection{Estimate UR via the Classifier}
\label{sec5.1}
By the chain rule for MI, the UR of feature $X_k$ in \eqref{URE} can be equivalently expressed as
\begin{align}
I(X_k;Y|\Omega \backslash X_k) = H(Y|\Omega \backslash X_k) - H(Y|\Omega). \label{EUR1}
\end{align}
The term $H(Y|\Omega)$ on the R.H.S of \eqref{EUR1} is constant for every candidate feature during the selection process. Therefore, boosting UR is equivalent to boosting $H(Y|\Omega \backslash X_k)$, which is a function of the likelihood $p(Y|\Omega \backslash X_{k})$ and this likelihood can be estimated using a classifier $\mathbb{Q}$ with parameter $\theta$. Assuming all samples are i.i.d, the estimated UR of feature $X_k$ can be rewritten as
\begin{align}
\hspace{-2mm} \text{UR} (X_k) \equiv H(Y|\Omega \backslash X_k) & = \mathbb{E} \left\{ \log{p(Y|\Omega \backslash X_k)} \right\}, \\
& \approx \frac{1}{N}\sum_{i=1}^{N} \log \mathbb{Q}(y^i|(\Omega^i \backslash x_k^i), \theta),
\end{align}
where $N$ is the number of samples. 

The reason for estimating UR via a classifier is two-fold:

\begin{enumerate}[noitemsep,leftmargin=5mm, topsep=0pt]
\item It is understood that the estimation of high-dimensional MI is challenging and arguably suffers from the curse of dimensionality \cite{bellman1966dynamic}. Our approach provides an alternative way to estimate UR and the estimated UR will approach the true value as the number of samples $N$ grows given the classifier $\mathbb{Q}$ is a consistent estimator \cite{lehmann2006theory}. 

\item We note that different classifiers may treat the UR of features differently due to their working mechanisms and assumptions. For example, MI quantifies a non-linear relationship between random variables and this non-linear relationship may not be of much help to linear classifiers (e.g., SVM with linear kernel). Estimating UR via a classifier attempts to adapt UR to different classifiers, further improving the competitiveness of MRwMR-BUR for classification performance oriented tasks.
\end{enumerate}

\subsection{Experiment Setup}
\label{sec5.2}
We now compare the performance of MRwMR-BUR using the classifier based approach to estimate UR to that of using the KSG estimator via KNN, SVM and RF. We highlight that the UR is estimated using the classifier being tested and all estimated URs are first normalized from 0 to 1 using min-max normalization. Furthermore, estimating UR in such a manner changes MRwMR-BUR based algorithms to a classifier dependent method. Therefore, in addition to the performance of MRwMR-BUR using the KSG estimator, 
three popular classifier dependent feature selection methods: (i) Recursive Feature Elimination (RFE) \cite{guyon2002gene}, (ii) Backward Feature Elimination (BFE) \cite{kohavi1997wrappers} and (iii) Forward Feature Selection (FFS) \cite{marcano2010feature}, are also used for performance evaluation.

The average test accuracy $\pm$ standard deviation over 20 runs and the corresponding number of features required (i.e., the value in the parentheses) are shown in Table \ref{tab:rf_2} with $\beta = 0.1$. The CLF named algorithms estimate UR via the classifier based approach. Since KNN is a non-parametric classifier contains no parameters related to the feature importance, RFE is not applicable to KNN. Therefore, the results of RFE on KNN is not shown.

\subsection{Performance Evaluation}
\label{sec5.3}
{\bf (1) MRwMR-BUR-CLF VS MRwMR. }In Table \ref{tab:rf_2}, we find that the classifier based approach further improves the performance of MRwMR based algorithms. For example, the accuracy of MIM on the Madelon dataset using RF is increased from 71.5\% to 74.2\% (compare row 1 in Table \ref{tab:rf} to row 1 in Table \ref{tab:rf_2}), with an improvement of 3.8\%. Similar trends can be observed using other classifiers. For example, the accuracy of MIM using KNN on the Madelon dataset is increased from 73.0\% to 77.0\% (compare row 11 in Table \ref{tab:rf} to row 9 in Table \ref{tab:rf_2}), with an improvement of 5.5\%. Note that the improvement is much higher than  MRwMR-BUR-KSG, which improves the accuracy of MIM from 73.0\% to 74.5\% (compare row 11 to row 12 in Table \ref{tab:rf}). This verifies our goal of adapting UR to different classifiers so as to further improve the performance of MRwMR-BUR based algorithms.



{\bf (2) MRwMR-BUR-CLF VS Classified Dependent Feature Selection Methods. }As compared to the classifier dependent methods, MRwMR-BUR-CLF also obtains a better performance. For example, the accuracy of GSA-BUR-CLF on the Colon dataset using KNN is 85.4\%, which is 1.5\% higher than BFE. Similarly, the accuracy of mRMR-BUR-CLF on the Isolet dataset using KNN is 1.8\% higher than FFS.



\subsection{MRwMR-BUR-KSG VS MRwMR-BUR-CLF}
Three factors are usually considered when choosing a feature selection method, i.e., (i) classification performance, (ii) feature interpretability and (iii) classifier generalization. Depending on the task being addressed, we need to apply different variants of MRwMR-BUR. For the classification performance oriented task, MRwMR-BUR-CLF should be used as it tends to provide better classification performance than MRwMR-BUR-KSG, as well as classifier dependent feature selection methods. If 
either of the other two factors (feature interpretability, classifier generalization) also needs to be considered, we recommend using MRwMR-BUR-KSG. This is because MRwMR-BUR-KSG is classifier independent and the selected feature subset is not tied to the bias of classifier. As a result, the feature subset selected by MRwMR-BUR-KSG does not vary from classifier to classifier, and thus is generally more accurate in terms of feature interpretability. Furthermore, the feature subset is selected from an information-theoretic perspective and tends to generalize well to various classifiers, since the mechanisms of most classifiers are closely related to information theory.


\begin{table*}[!t]
\setlength\tabcolsep{5.5pt}
{\renewcommand{\arraystretch}{1.58}
\centering
\begin{tabular}{l|cccccccc} \hline
\hline                       
 & Colon & Sonar & Madelon & Leukemia & Isolet   & Gas sensor\\ \hline
  
\multicolumn{7}{c}{\bf \em Random Forest (RF)} \\ \hline

1) MIM-BUR-CLF      & 86.3 $\pm$ 2.9 (18)    & 80.3 $\pm$ 1.1 (33)   & 74.2 $\pm$ 0.8 (55)  & 98.4 $\pm$ 1.4 (37)  & 87.5 $\pm$ 1.0 (119)    & 99.47 $\pm$ 0.06 (70) \\ 

2) mRMR-BUR-CLF &  {\small \textbf{88.7 $\pm$ 1.6 (30)}}  &  81.4 $\pm$ 1.0 (35)   &   73.1 $\pm$ 0.5 (54)  & 98.8 $\pm$ 1.6 (50) & 88.3 $\pm$ 0.3 (127)   & 99.49 $\pm$ 0.03 (65) \\

3) JMI-BUR-CLF      &  75.5 $\pm$ 1.1 (34)   &  81.4 $\pm$ 2.0 (35)   &  72.3 $\pm$ 0.9 (58)  & {\small \textbf{99.2 $\pm$ 0.8 (62)}}  &87.9 $\pm$ 0.4 (119)  &  99.48 $\pm$ 0.03 (70) \\

4) JMIM-BUR-CLF   &  76.2 $\pm$ 1.9 (17)  &  81.0 $\pm$ 0.9 (43)  & 73.4 $\pm$ 0.3 (54)   &  99.2 $\pm$ 1.1 (70)  & 85.9 $\pm$ 0.8 (107)   & 99.49 $\pm$ 0.04 (70)  \\  

5) GSA-BUR-CLF    &  88.0 $\pm$ 0.9 (20)  & {\small \textbf{81.9 $\pm$ 1.4 (33)}}  &  74.5 $\pm$ 0.3 (55)   &  99.0 $\pm$ 1.4 (59)  &   {\small \textbf{88.8 $\pm$ 0.4 (96)}}   &  {\small \textbf{99.50 $\pm$ 0.06 (81)}}        \\\hline
6) RFE & 87.1 $\pm$ 1.4 (23) & 80.9 $\pm$ 1.5 (31) & {\small \textbf{74.6 $\pm$ 0.5 (51)}} & 98.8 $\pm$ 0.7 (42) & 88.1 $\pm$ 0.5 (102) & 99.49 $\pm$ 0.05 (68) \\ 
7) FFS & 87.7 $\pm$ 0.8 (25) & 81.2 $\pm$ 1.2 (36) & 74.4 $\pm$ 0.3 (54) & 98.9 $\pm$ 0.6 (48) & 87.9 $\pm$ 0.5 (107) & 99.43 $\pm$ 0.04 (63) \\ 
8) BFE & 86.9 $\pm$ 1.1 (27) & 80.8 $\pm$ 0.7 (39) & 74.5 $\pm$ 0.3 (49) & 98.7 $\pm$ 0.9 (53) & 88.3 $\pm$ 0.5 (111) & 99.46 $\pm$ 0.03 (72) \\ \hline
 \multicolumn{7}{c}{\bf \em K-Nearest Neighbors (KNN)} \\ \hline

 9) {MIM-BUR-CLF}            &   86.3 $\pm$ 1.8 (41)   &    84.6 $\pm$ 0.7 (41)    & 77.0 $\pm$ 0.6 (63)   &  97.7 $\pm$ 0.4 (31)    &  80.1 $\pm$ 0.8 (123) & 99.09 $\pm$ 0.05 (78)              \\ 
  
 10) {mRMR-BUR-CLF}       &   86.3 $\pm$ 1.9 (29)    & 84.8 $\pm$ 1.4 (30)     &  76.3 $\pm$ 1.4 (65)   &  98.5 $\pm$ 0.8 (46)   &  \small \textbf{82.6 $\pm$ 1.1 (107)}  & 99.17 $\pm$ 0.06 (71) \\ 

 11) {JMI-BUR-CLF}            &   71.2 $\pm$ 1.6 (38)    & 84.5 $\pm$ 0.9 (46)      & \small \textbf{79.9 $\pm$ 1.0 (57)}   &   98.8 $\pm$ 0.7 (34)    & 80.8 $\pm$ 1.3 (119)  &  \small \textbf{99.28 $\pm$ 0.06 (76)}  \\ 

 12) {JMIM-BUR-CLF}         &   77.2 $\pm$ 1.4 (51)     & 84.1 $\pm$ 0.9 (40)    &  79.3 $\pm$ 0.6 (52)   &   \small \textbf{99.2 $\pm$ 0.7 (28)}   &  80.1 $\pm$ 1.0 (121)  & 98.78 $\pm$ 0.03 (75)  \\ 
 
 13) {GSA-BUR-CLF}          &   {\small \textbf{88.3 $\pm$ 1.5 (50)}}      &   {\small \textbf{85.4 $\pm$ 1.0 (44)}}   & 78.4 $\pm$ 1.1 (40)   &  98.4 $\pm$ 0.9 (51)     &  81.0 $\pm$ 0.9 (107)   &  99.2 $\pm$ 0.05 (65) \\ \hline
14) FFS & 86.8 $\pm$ 1.1 (33) & 83.7 $\pm$ 1.2 (41) & 78.8 $\pm$ 0.8 (57) & 98.9 $\pm$ 0.6 (32) & 81.3 $\pm$ 0.8 (105) & 99.12 $\pm$ 0.04 (62) \\
15) BFE & 87.4 $\pm$ 1.5 (37) & 84.1 $\pm$ 0.8 (44) & 79.2 $\pm$ 1.0 (61) & 98.5 $\pm$ 0.9 (35) & 81.1 $\pm$ 0.7 (109) & 99.07 $\pm$ 0.07 (71) \\ \hline

 \multicolumn{7}{c}{\bf \em Support Vector Machine (SVM)} \\ \hline

 16) {MIM-BUR-CLF}          &   82.4 $\pm$ 2.3 (59)  &  74.1 $\pm$ 0.6 (45)  & 62.8 $\pm$ 0.6 (40)  &  98.1 $\pm$ 0.9 (78)  & 89.1 $\pm$ 0.5 (104)  &  97.1 $\pm$ 0.04 (87)  \\ 

 17) {mRMR-BUR-CLF}      &  {\small \textbf{84.2 $\pm$ 0.8 (34)}}   & {\small \textbf{75.5 $\pm$ 0.8 (45)}}  & 62.1 $\pm$ 0.7 (43)  &   {\small \textbf{98.7 $\pm$ 0.7 (59)}}  & 89.9 $\pm$ 0.7 (120)  & 96.4 $\pm$ 0.05 (83)    \\ 

 18) {JMI-BUR-CLF}           & 73.3 $\pm$ 1.6 (20)   &  75.4 $\pm$ 1.0 (39) & 62.2 $\pm$ 0.6 (34)  &   98.0 $\pm$ 0.9 (75)  &   {\small \textbf{90.1 $\pm$ 1.3 (117)}}  &  97.1 $\pm$ 0.05 (92)    \\ 

 19) {JMIM-BUR-CLF}        &   79.5 $\pm$ 1.3 (26)  & 73.7 $\pm$ 1.4 (35)  & 62.5 $\pm$ 0.6 (42)  &  98.3 $\pm$ 0.6 (70) &   89.4 $\pm$ 1.0 (129)   &   {\small \textbf{97.3 $\pm$ 0.09 (93)}}    \\ 
       
 20) {GSA-BUR-CLF}         &   76.2 $\pm$ 1.1 (29)  & 74.1 $\pm$ 1.1 (43)  & 63.4 $\pm$ 0.5 (36)  &  98.5 $\pm$ 0.9 (66) &   90.1 $\pm$ 1.2 (122)   &   96.5 $\pm$ 0.03 (86)    \\ \hline
21) RFE & 83.8 $\pm$ 0.9 (33) & 74.7 $\pm$ 1.5 (41) & 63.5 $\pm$ 0.9 (35) & 98.6 $\pm$ 0.6 (68) & 89.7 $\pm$ 0.9 (118) & 96.1 $\pm$ 0.1 (85)\\  
22) FFS & 83.1 $\pm$ 1.2 (31) & 74.5 $\pm$ 1.2 (36) & 63.8 $\pm$ 0.7 (37) & 98.5 $\pm$ 0.7 (62) & 89.4 $\pm$ 0.5 (109) & 96.4 $\pm$ 0.07 (82) \\
23) BFE & 83.4 $\pm$ 1.1 (35) & 74.8 $\pm$ 1.0 (39) & {\small \textbf{63.9 $\pm$ 0.6 (38)}} & 98.6 $\pm$ 0.8 (66) & 89.6 $\pm$ 0.8 (119) & 96.3 $\pm$ 0.06 (88) \\ \hline \hline
\end{tabular}
\captionsetup{justification=raggedright,skip=5pt}
\vspace{-1mm}
\caption{Performance of MRwMR-BUR-CLF based algorithms (with $\beta$ = 0.1) and three popular classifier dependent feature selection methods (RFE, FFS and BFE). The results are presented as: average test accuracy (\%) $\pm$ standard deviation over 20 runs (number of features required). The results are shown for RF (rows 1 - 8), KNN (rows 9 - 15) and SVM (rows 16 - 23). The bold indicates the best performance over all methods studied for a given classifier.}
\label{tab:rf_2}
}
\end{table*}

\section{Reflections}
\label{reflections}
In this section, we conclude the paper by presenting some reflections and suggestions for future work.

\textbf{(1) Advantage of MRwMR-BUR over MRwMR.} In this paper, we propose MRwMR-BUR as a new criterion to design MIBFS algorithms. The MRwMR-BUR criterion equips the existing MRwMR criterion with a mechanism to explicitly target the features with UR. The MRwMR based algorithms can be easily modified to the corresponding MRwMR-BUR form, without increasing the complexity. The experimental results suggest that the MRwMR-BUR helps to better achieve the goal of MIBFS by selecting fewer number of redundant features, leading to better classification performance, interpretability and generalization. 


\textbf{(2) Should We always Select Features with UR First?} The results in Section \ref{sec3.1} show that the minimal feature subset $S^*$ in \eqref{NewP} must include all features with UR. Does this mean that we should only select features with UR first? The answer is no, especially when we have a feature budget, i.e., the number of features that can be selected is limited. This is because features with UR (e.g., $I(X_k;Y | \Omega \backslash X_k)>$0) may not contribute to higher relevance in the short term conditioned on $S \subseteq\Omega \backslash X_k$ (i.e., $I(X_k;Y|S)=0$) as reducing conditional features (from $\Omega \backslash X_k$ to $S \subseteq\Omega \backslash X_k$) may decrease the conditional MI.

\textbf{(3) Optimal Value of $\beta$.} The performance of MRwMR-BUR based algorithms relies on the value of $\beta$, which balances the original objective of maximizing relevance with prioritizing features with UR. From extensive experiments, we have consistently found that $\beta$ = 0.1 is a good choice to balance UR and relevance. Alternatively, $\beta$ can be thought of as a hyper-parameter and tuned via a validation dataset. The theoretical determination of the optimal value $\beta$ is clearly worth deeper thought. This could help to further minimize redundancy by prioritizing more important features.

\textbf{(4) Potentially Better MIBFS Algorithms.} In this paper, we are not proposing a new MIBFS algorithm. Instead, we explore a new criterion for MIBFS algorithms that incorporate UR into the objective. Our results demonstrate that the MRwMR-BUR criterion has superior performance over the existing MRwMR criterion. We believe this new insight can inspire better MIBFS algorithms that optimally use UR.

\newpage
\bibliographystyle{IEEEtran}
\bibliography{bare_jrnl_compsoc}

\begin{IEEEbiography}[{\includegraphics[width=1in,height=1.25in,clip,keepaspectratio]{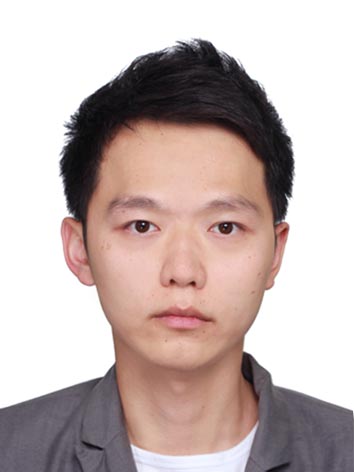}}]%
{Liu Shiyu}
received his B.Eng degree in Electrical Engineering in 2018 from National University of Singapore. Since 2018, he is a PhD candidate at National University of Singapore. His major research interests include feature engineering, time series data forecasting and neural network compression. 
\end{IEEEbiography}

\begin{IEEEbiography}[{\includegraphics[width=1in,height=1.6in,clip,keepaspectratio]{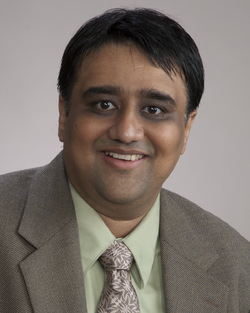}}]%
{Mehul Motani}
received the B.E. degree from Cooper Union, New York, NY, the M.S. degree from Syracuse University, Syracuse, NY, and the Ph.D. degree from Cornell University, Ithaca, NY, all in Electrical and Computer Engineering. Dr. Motani is currently an Associate Professor in the Electrical and Computer Engineering Department at the National University of Singapore (NUS) and a Visiting Research Collaborator at Princeton University.His research interests include information and coding theory, machine learning, biomedical informatics, wireless and sensor networks, and the Internet-of-Things.
\end{IEEEbiography}

\end{document}